\documentclass[11pt,a4paper]{article}
\usepackage{graphicx}
\usepackage{xspace} 

\usepackage[final]{acl}
\usepackage{tabularx}
\usepackage{ragged2e}
\usepackage{times}
\usepackage{latexsym}
\usepackage{enumitem}
\usepackage{hyperref}
\usepackage{amsmath}
\usepackage[table]{xcolor}
\usepackage[T1]{fontenc}
\usepackage[utf8]{inputenc}

\usepackage{microtype}

\usepackage{inconsolata}

\usepackage{graphicx}
\usepackage{booktabs}
\newcommand{\huggingface}{\raisebox{-2pt}{\includegraphics[height=1.3em]{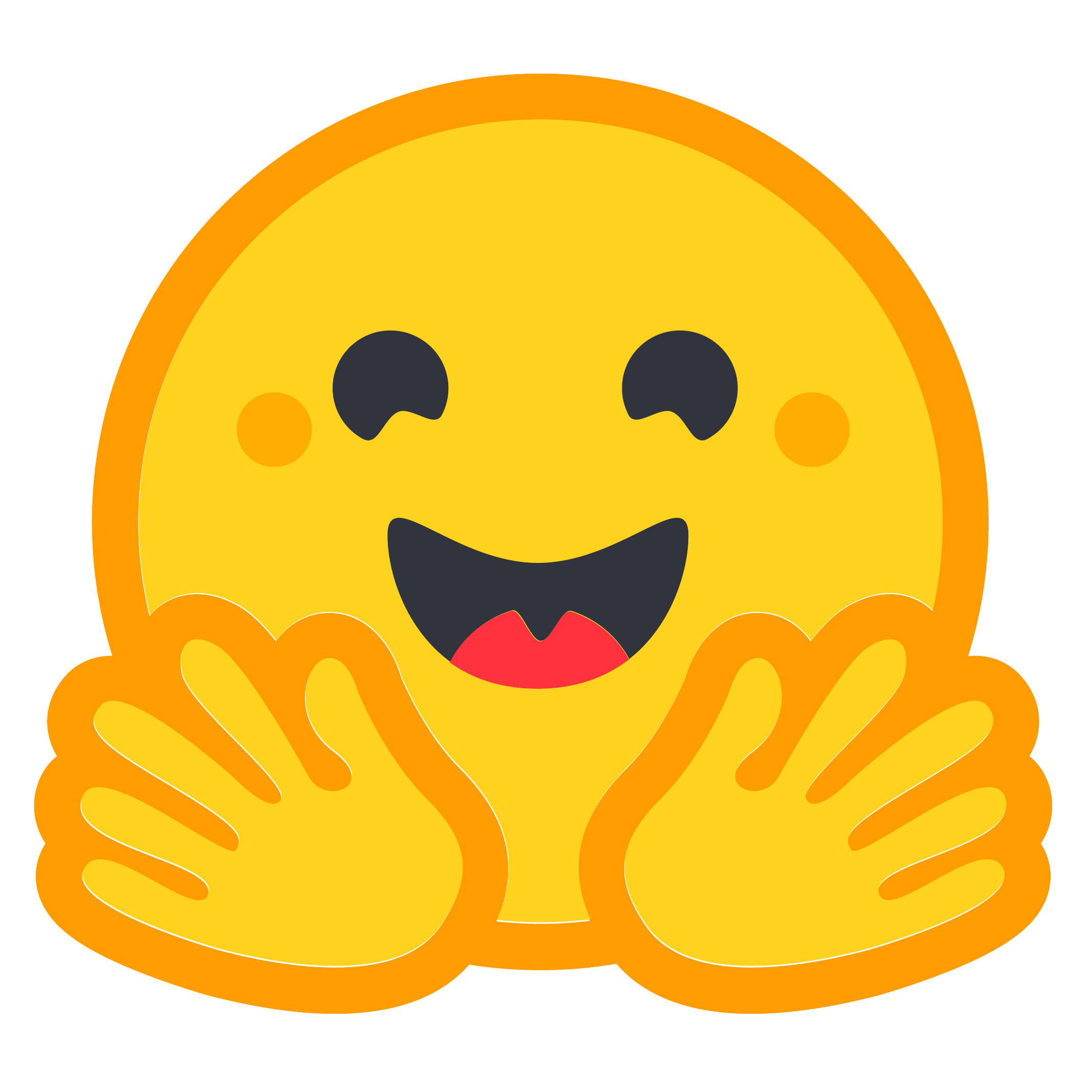}}\xspace}
\newcommand{\github}{\raisebox{-2pt}{\includegraphics[height=1.3em]{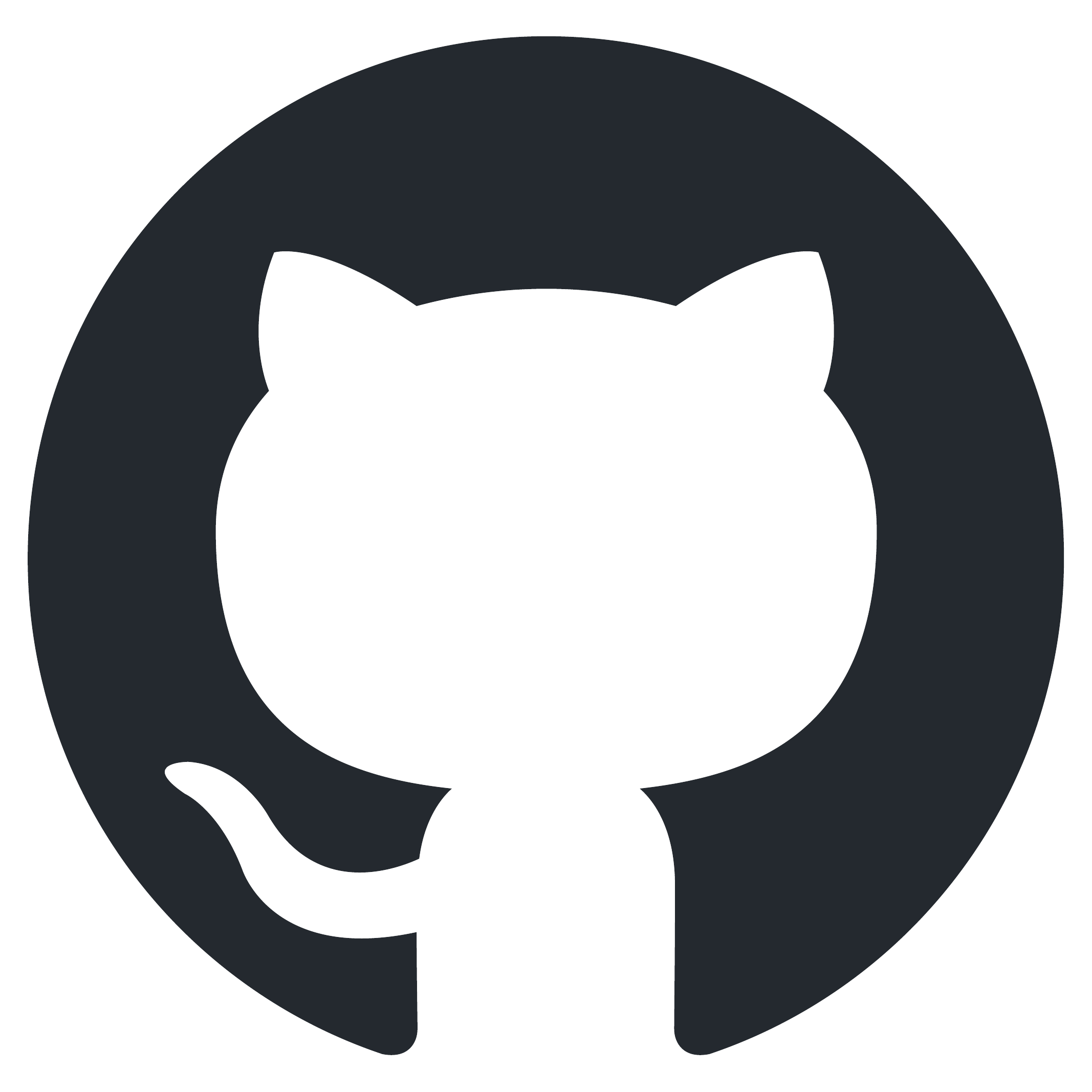}}\xspace}

\usepackage[table]{xcolor}
\definecolor{ctxgray}{gray}{0.95}
\definecolor{diffneg}{RGB}{180, 50, 50}
\definecolor{diffpos}{RGB}{50, 150, 50}
\definecolor{arrowred}{RGB}{180, 50, 50}
\definecolor{arrowgreen}{RGB}{50, 150, 50}
\definecolor{ctxblue}{RGB}{245, 245, 255}

\newcommand{\diffdown}[1]{\textcolor{arrowred}{$\downarrow$ #1}}
\newcommand{\diffup}[1]{\textcolor{arrowgreen}{$\uparrow$ #1}}

\author{
    \textbf{Yongan Yu}$^{M,Q\dagger}$,
    \textbf{Shantam Raj}$^{Z\dagger}$,
    \textbf{Jingwei Ni}$^{Z,E*}$\\
    \textbf{Ario Saeid Vaghefi}$^{Z}$, 
    \textbf{Dominik Stammbach}$^{P}$,
    \textbf{Markus Leippold}$^{Z*}$\\
    $^{M}$McGill University
    \quad
    $^{Z}$University of Zürich\\
    $^{E}$ETH Zürich
    \quad
    $^{P}$Princeton University\\
    $^{Q}$Mila -- Quebec Artificial Intelligence Institute\\
    \texttt{yongan.yu@mail.mcgill.ca}, 
    \texttt{shantam.raj@uzh.ch}\\
\vspace{.5em}\huggingface \href{https://huggingface.co/CMB-ClimateModernBERT}{Models}
\hspace{0.6em}
\vspace{.5em}\github \href{https://github.com/ClimateModernBERT/ClimateModernBERT}{Code}
}

\title{Climate-ModernBERT: Revisiting Corpus Composition for Domain-Adaptive Continued Pretraining}

\begin{document}
\maketitle
\begin{abstract}
Natural Language Processing (NLP) in the climate domain requires models to process heterogeneous text sources, including scientific literature, policy disclosures, and synthetic reports. However, how to effectively combine diverse domain corpora during continued pretraining (CPT) remains underexplored.
We introduce \textsc{Climate-ModernBERT}, a family of climate-adapted encoder models obtained through continued pretraining of ModernBERT-Base on three climate corpora: academic climate text, climate-filtered web data, and synthetic climate documents.
We systematically compare joint continued pretraining on corpus mixtures with parameter-space merging of independently specialized checkpoints. 
Across nine climate NLP benchmarks, our best model achieves 76.3 average F$_1$, improving significantly over a vanilla ModernBERT baseline by 2.8 points. 
Within the climate NLP setting, the results show that academic climate corpora provide the strongest adaptation signal among the evaluated sources, while parameter-space merging improves over joint multi-source training and better preserves complementary information from heterogeneous climate corpora.
We release all \textsc{Climate-ModernBERT} variants and training checkpoints
% \footnote{Github Repo: \url{https://github.com/ClimateModernBERT/ClimateModernBERT}}
to support future research in climate NLP and domain-adaptive pretraining.
\end{abstract}

\renewcommand{\thefootnote}{\fnsymbol{footnote}}
\footnotetext{$\dagger$ Equal contribution;
\(\ast\) Work conducted under the supervision of these authors.
}
\renewcommand{\thefootnote}{\arabic{footnote}}

\section{Introduction}
The rapid escalation of the global climate crisis increases the need for advanced NLP tools that can parse, synthesize, and verify large corpora of environmental literature and policy disclosures \citep{hershcovich2022towards}.
Various efforts in the NLP community aim to support the automatic analysis of climate-related text through tasks such as climate topic detection \citep{varini2020climatext, yu2024climatebug}, climate risk classification \citep{bingler2024cheap}, and question answering over climate knowledge \citep{spokoyny2023towards}. 
These tools enable practitioners in the climate domain to extract insights from rapidly growing collections of climate reports, scientific publications, and regulatory corpus.

Yet building strong NLP models for this domain remains challenging in practice. Climate text resources are highly heterogeneous \citep{shi2023detecting}, spanning informal web discourse, scientific literature, and unstructured policy documents, while evaluation benchmarks cover diverse NLP tasks across short and long documents with specialized terminology and concepts. Climate practitioners must therefore operate over noisy and stylistically inconsistent corpora, making it difficult to train a single domain-adapted model that performs reliably across diverse downstream applications. Recent advances in encoder architectures provide a promising direction for addressing these challenges. ModernBERT \citep{warner2025smarter} introduces a compact yet powerful encoder model with strong performance across natural language understanding tasks. Compared with large generative decoders, encoder-only models offer a favorable balance between performance and efficiency, making them suitable for large-scale climate analysis pipelines. They support non-generative tasks such as classification, retrieval, and information extraction, and serve as key components in retrieval systems and retrieval-augmented generation pipelines over large document collections \citep{lewis2020retrieval}. These properties make them well-suited for climate applications requiring long-document processing and nuanced semantic understanding across heterogeneous text sources.

\begin{figure*}[t!]
    \centering
    \includegraphics[width=0.95\textwidth]{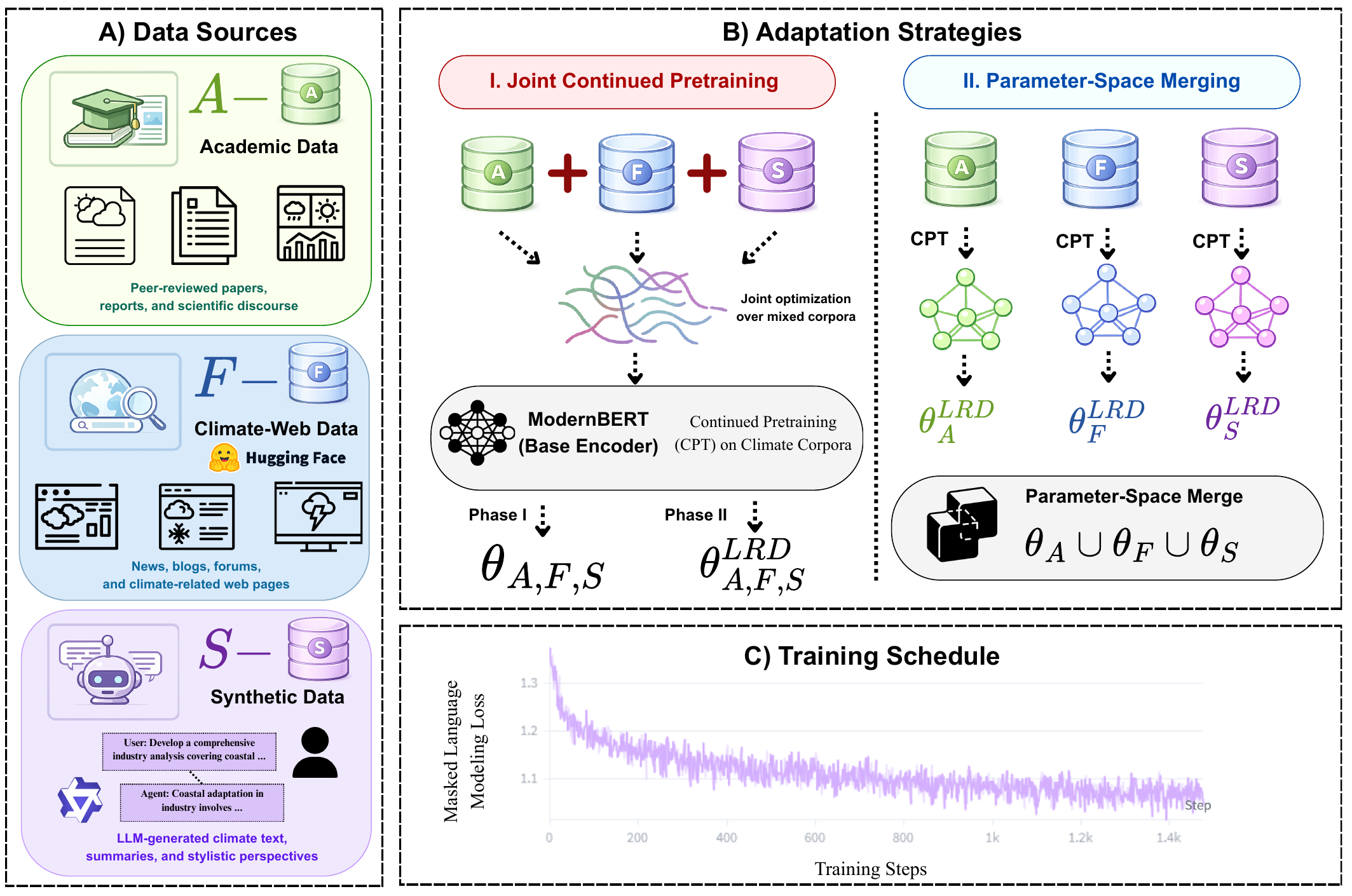}
    \caption{
Overview of the \textsc{Climate-ModernBERT} adaptation framework. 
(A) Three primary climate-domain corpora used for continued pretraining: academic climate text ($\mathcal{A}$), climate-filtered web data ($\mathcal{F}$), and synthetic climate text ($\mathcal{S}$). 
(B) Two adaptation strategies studied in this work: joint continued pretraining on mixed corpora and parameter-space merging of independently specialized checkpoints. 
(C) Following the ModernBERT training schedule, consisting of context extension followed by learning-rate decay specialization. Context-extension-only checkpoints are additionally evaluated as an ablation.
}
\label{full pipeline}
\vspace{-2ex}
\end{figure*}

Evidence from other scientific domains further supports the benefits of adapting ModernBERT through domain-specific pretraining. Models such as Bioclinical ModernBERT \citep{sounack2025bioclinical}, Clinical ModernBERT \citep{lee2025clinical}, and Legal ModernBERT \citep{stammbach2026legal} show that continued pretraining on large-scale domain corpora consistently improves performance on specialized downstream tasks. These results suggest that a similar strategy could benefit climate NLP systems. However, the role of corpus composition in domain adaptation remains poorly understood, despite growing evidence that different in-domain data sources can lead to substantially different downstream outcomes \citep{chalkidis2020legal, longpre2024pretrainer}.

As such, an important question remains largely underexplored in climate NLP: \textbf{how should heterogeneous text sources be combined during continued pretraining to adapt modern encoder architectures effectively?} 
We study two complementary aspects of this problem: (1) the composition of the training corpus and the mechanism used to integrate multiple sources, and (2) either through joint continued pretraining or parameter-space merging of independently specialized checkpoints.
More broadly, this case study provides empirical insights into corpus composition and adaptation strategies for climate-domain encoder adaptation, while motivating future investigations in other specialized NLP domains.

We introduce \textbf{\textsc{Climate-ModernBERT}}, a family of climate-adapted encoder models obtained through ModernBERT's continued pretraining procedure on heterogeneous climate corpora. Figure~\ref{full pipeline} provides an overview of the corpora, adaptation strategies, and two-stage training schedule. Our contributions are threefold:
\begin{itemize}[nosep]
    \item We introduce \textsc{Climate-ModernBERT}, a suite of climate-adapted ModernBERT encoders trained on three primary climate corpora: academic climate text, climate-filtered web data, and synthetic climate documents.
    \item We conduct a large-scale empirical study of corpus composition for climate-domain encoder adaptation, evaluating 21 climate-adapted model variants across nine climate NLP benchmarks, yielding over 810 fine-tuning runs.
    \item We compare joint continued pretraining and parameter-space merging under matched data coverage, showing that curated academic climate corpora provide the strongest adaptation signal and that parameter-space merging outperforms joint multi-source training.
\end{itemize}

\section{Related Work}

\subsection{Encoder-only Models in Climate NLP}
Encoder-based language models remain a core component of climate NLP systems, supporting tasks such as climate-risk assessment \citep{diggelmann2020climate}, sustainability disclosure analysis \citep{bingler2024cheap}, information retrieval \citep{schimanski2024climretrieve}, and scientific literature mining \citep{manivannan2024climaqa}. 
Domain-adapted encoders such as ClimateBERT \citep{webersinke2021climatebert} demonstrate strong performance across climate-specific classification and retrieval tasks and remain widely adopted in climate NLP pipelines. 
Recent evaluations further show that fine-tuned encoders, including ClimateBERT-based models, often match or exceed the zero-shot performance of larger decoder-based models on supervised climate NLP tasks when sufficient labeled data are available \citep{trajanov2023comparing, bucher2024fine}.

\subsection{Continued Pretraining for Domain Adaptation}
Continued pretraining, also known as domain-adaptive pretraining, is a widely adopted approach for adapting pretrained language models to specialized domains. Prior work demonstrates consistent gains from CPT in biomedical \citep{sounack2025bioclinical}, clinical \citep{lee2025clinical}, legal \citep{stammbach2026legal}, and spatial \citep{singh2025spatial} settings. Existing studies, however, primarily focus on the benefits of domain adaptation itself or the impact of factors such as training duration, optimization, and data selection \citep{chen2025towards, shi2025continual}. 
This leaves an open question of how multiple data sources should be combined during continued pretraining \citep{chen2025towards}. 
This question is particularly challenging in climate NLP, where training data span scientific literature, policy documents, web discourse, and synthetic text with substantially different distributions.

\subsection{Parameter-Space Model Merging}

When multiple data sources are available, the standard approach is to combine them during training through joint continued pretraining. An alternative is to adapt separate checkpoints for each source and combine them afterward in parameter space. Early work shows that simple weight averaging can improve generalization when models share a common initialization \citep{ wortsman2022model}. 
Subsequent methods, including Task Arithmetic \citep{ilharco2023editing}, TIES-Merging \citep{yadav2023ties}, and DARE \citep{yu2024language}, introduce mechanisms for combining parameter updates from independently trained models, typically in multi-task learning settings where each model is fine-tuned on a distinct downstream task. Their use as a strategy for integrating heterogeneous domain corpora during continued pretraining remains underexplored. We investigate this setting in climate NLP by comparing parameter-space merging against joint continued pretraining under a matched data budget.

\section{\textsc{Climate-ModernBERTs}}

\subsection{Data Curation}
\label{sec:data}
High-quality and diverse training corpora are essential for effective language modeling \citep{longpre2024pretrainer}, particularly in climate NLP, where language spans scientific, policy, and public discourse. 
To study how corpus composition influences continued pretraining, we construct a 6.42B-token climate corpus from three primary sources: (1) academic climate data, (2) climate-filtered web data, and (3) synthetic climate text. Table~\ref{tab:corpus-summary} summarizes the high-level composition. 

\begin{table}[t]
\centering
\small
\setlength{\tabcolsep}{4pt}
\begin{tabular}{l l r r}
\toprule
\textbf{Corpus} & \textbf{Format} & \textbf{\#Docs} & \textbf{Tokens} \\
\midrule
$\mathcal{A}$ (Academic)   & XML\,/\,CSV\,/\,PDF & $\sim$5\,M   & $\sim$1.28\,B \\
$\mathcal{F}$ (Web)        & Parquet             & 2.59\,M       & $\sim$5\,B \\
$\mathcal{S}$ (Synthetic)  & JSONL               & $\sim$20\,K  & $\sim$0.14\,B \\
\midrule
\textbf{Total}             & ---                 & ---          & $\sim$6.42\,B \\
\bottomrule
\end{tabular}
\caption{High-level composition of the climate pretraining corpus across the three sources $\mathcal{A}$, $\mathcal{F}$, and $\mathcal{S}$.}
\vspace{-3ex}
\label{tab:corpus-summary}
\end{table}

\paragraph{Academic Climate Data ($\mathcal{A}$)}
The academic corpus aggregates four complementary sources of climate-related text: (i) peer-reviewed journal articles spanning climate science, earth systems, and energy economics; (ii) the \textit{ClimateNews} archive containing climate-related news articles from 2000--2022; (iii) climate-focused arXiv preprints sampled from categories including \texttt{cs}, \texttt{physics.ao-ph}, \texttt{econ}, and \texttt{q-fin.RM}; and (iv) further included reports from the Intergovernmental Panel on Climate Change (IPCC)\footnote{\url{https://www.ipcc.ch/data/}} and climate handbooks \citep{pachauri2014climate}. 
We treat these sources as a unified academic corpus due to their substantial overlap in vocabulary, citation conventions, and discourse structure. Document-level layout (e.g., sections, paragraphs, and captions) is preserved whenever possible to support long-context training with ModernBERT's 8,192-token context window. After applying deduplication and decontamination preprocessing pipeline (details are described in Appendix~\ref{app:preprocessing}), we arrive at a corpus of 1.28B tokens. A detailed per-source breakdown is reported in Appendix~\ref{app:academic-data}.

\paragraph{Web Data ($\mathcal{F}$)}
The web corpus is derived from FineWeb-Edu \citep{penedo2024fineweb}, a large-scale high-quality web dataset containing diverse educational and informational content. 
We extract climate-relevant documents using a two-stage filtering pipeline. We apply a high-recall keyword filter built from 166 climate-related terms followed by deduplication (detailed in Appendix~\ref{app:web-keywords}), yielding 2.59M candidate documents. 
% Second, we improve precision using a FastText classifier \citep{bojanowski2017enriching} trained on 10K manually annotated samples labeled with GPT-4o. 
% Retaining documents with calibrated climate probability $p_{\text{climate}} \geq 0.5$ results in a final subset of approximately 640K documents. 
% After preprocessing and deduplication, the filtered web corpus contains approximately 0.8B tokens.

\paragraph{Synthetic Climate Data ($\mathcal{S}$)}
To improve coverage of underrepresented climate subdomains \citep{marwala2023use}, we generate synthetic climate text conditioned on seed excerpts drawn from six in-domain sources: five academic journal subsets from $\mathcal{A}$ (\textit{Climate Risk Management}, \textit{Environmental Science}, \textit{Renewable \& Sustainable Energy Reviews}, \textit{Urban Climate}, and \textit{Weather and Climate Extremes}) together with the \textit{ClimateNews} archive sampled across 22 years. For each seed document, we extract an 800-character passage and prompt \texttt{Qwen3.5-122B-A10B} \citep{qwen3.5} to generate text in three climate communication perspectives inspired by \citet{schafer2021climate}: \textit{public awareness}, \textit{industry perspective}, and \textit{environmental journalism}. Generation details, prompt templates, and seed statistics are provided in Appendix~\ref{app:synthetic-prompts}. 

\subsection{Training Schedule}

We adapt ModernBERT-base \citep{warner2025smarter} following its original continued pretraining procedure. Specifically, we apply context extension (CX) followed by learning-rate decay (LRD) specialization, where the resulting CX+LRD checkpoints serve as the final climate-adapted models used throughout our main experiments. We additionally evaluate CX-only checkpoints as a case study to isolate the contribution of LRD specialization.

\paragraph{Initialization.}
Following prior domain-adaptive ModernBERT studies \citep{sounack2025bioclinical,lee2025clinical}, we initialize from the public ModernBERT-base context-extension stage checkpoint. This checkpoint preserves the stable optimization landscape, which facilitates continued training on new domain data. In contrast, resuming from a post-LRD checkpoint would require reintroducing warm-up \citep{ash2020warm} to recover from the near-zero learning rate reached during decay \citep{hagele2024scaling}.

\paragraph{Context Extension.}
We retain the context-extension checkpoints as an ablation to examine the contribution of LRD specialization. CX training uses a constant learning rate of $3e-4$, global batch size of $576$, sequence length of $8{,}192$, and MLM masking rate of $30\%$. We optimize with Decoupled StableAdamW \citep{wortsman2023stable} in BF16 mixed precision for three epochs.

\paragraph{LRD specialization.}
Following the original ModernBERT training procedure, we further specialize each CX checkpoint using the $1-\sqrt{t}$ learning-rate decay schedule \citep{warner2025smarter}. 
We use an initial learning rate of $3e-4$ and a final learning-rate factor $\alpha_f=1e-3$ for three additional epochs, while keeping all other hyperparameters unchanged.
The resulting CX+LRD checkpoints serve as the final climate-adapted models and are used for all subsequent corpus-composition and model-merging experiments described in \S\ref{sec:models}.

\paragraph{Model Architecture and Computing.}

Every variant uses the ModernBERT-base architecture ($150$M parameters, $22$ transformer layers, hidden size $768$, $12$ attention heads, and a $50{,}368$-token vocabulary), with RoPE positional embedding and alternating $128$-token-window local and global attention with a global block every three layers \citep{wu2025emergence}. 
Training runs on $4{\times}$ NVIDIA A100 GPUs using the MosaicML Composer framework \citep{warner2025smarter}. Checkpoints are saved every $3{,}000$ steps during CX training and every $1{,}000$ steps during LRD specialization. 
Final checkpoints are converted to the HuggingFace Transformers FlexBERT format for release. Complete hyperparameter and architecture specifications are deferred to Appendix~\ref{app:training-setup}.

\section{Evaluation Setting}

We evaluate all climate-domain variants against two complementary baselines: the ModernBERT-base stable-phase checkpoint used for initialization and the original ClimateBERT model \citep{webersinke2021climatebert}, a widely adopted climate-domain encoder. The ModernBERT-base comparison isolates the effect of climate-domain continued pretraining from the original model capabilities, while ClimateBERT provides context against an established climate-adapted encoder. We fine-tune all models under the same downstream protocol to ensure a controlled comparison. We do not adopt the final publicly released ModernBERT checkpoint as the primary baseline, as it has already undergone full learning-rate decay on general-domain data, which would confound assessment of domain adaptation effects; for completeness, we still fine-tune that post-LRD checkpoint (results in Appendix~\ref{app:modernbert-full}).

\subsection{Evaluated Models}
\label{sec:models}

Our training design studies two aspects of climate adaptation: (i) the composition of the climate pretraining corpus and (ii) the contribution of LRD specialization through a CX-only ablation. Let the set of domain corpora introduced in \S\ref{sec:data} be $\mathcal{D}=\{\mathcal{A}, \mathcal{S}, \mathcal{F}\}$, where $\mathcal{A}$ denotes the academic climate corpus, $\mathcal{S}$ the synthetic climate corpus, and $\mathcal{F}$ the climate-filtered FineWeb-Edu subset.

For any non-empty subset $\mathcal{X}\subseteq\mathcal{D}$, we denote by $\theta_{\mathcal{X}}^{\textsc{lrd}}$ the final climate-adapted checkpoint obtained after CX followed by LRD specialization on the union of corpora in $\mathcal{X}$. We additionally denote by $\theta_{\mathcal{X}}$ the corresponding CX-only checkpoint used for ablation analysis. We further denote by $\theta_{\textit{base}}$ the stable-phase ModernBERT-base initialization from which all models are derived and against which all comparisons are made. 
% Throughout the paper, models are identified by the pair $(\mathcal{X}, \textsc{LRD})$. 

\paragraph{Joint-training variants.}
We train seven joint-training variants corresponding to different non-empty subsets of \[
\mathcal{D}:
\begin{array}{c}
\{\mathcal{A}\},\,
\{\mathcal{S}\},\,
\{\mathcal{F}\},\\[2pt]
\{\mathcal{A},\mathcal{S}\},\,
\{\mathcal{A},\mathcal{F}\},\,
\{\mathcal{S},\mathcal{F}\},\,
\{\mathcal{A},\mathcal{S},\mathcal{F}\}.
\end{array}
\]

\noindent For each subset $\mathcal{X}$, we train the final LRD-specialized checkpoint $\theta_{\mathcal{X}}^{\textsc{lrd}}$ used in our main experiments. The corresponding CX-only checkpoint $\theta_{\mathcal{X}}$ is retained for ablation analysis to evaluate the contribution of LRD specialization.

\paragraph{Model-merging variants.}
We further study parameter-space merging over independently trained checkpoints. Specifically, we take the three single-source LRD-specialized checkpoints $\theta_{\{c\}}^{\textsc{lrd}}$ with $c \in \{\mathcal{A}, \mathcal{S}, \mathcal{F}\}$ and combine their parameter deltas relative to $\theta_{\textit{base}}$ using four merging strategies: simple weight averaging ($\theta_{\textsc{Soup}}$), Task Arithmetic with scaling factors $\lambda \in \{0.5, 1.0\}$ ($\theta_{\textsc{Ta}(\lambda)}$), TIES-Merging with drop ratios $d \in \{0.5, 0.7\}$ ($\theta_{\textsc{Ties}(d)}$), and DARE-TIES with the same drop ratios ($\theta_{\textsc{Dare}(d)}$). 
This trajectory yields seven merged checkpoints whose training data correspond to the full corpus union $\{\mathcal{A}, \mathcal{S}, \mathcal{F}\}$, enabling direct comparison between joint training and post-hoc parameter merging under matched data coverage.

\begin{table*}[t]
\centering
\small
\setlength{\tabcolsep}{4pt}
\renewcommand{\arraystretch}{1.3}
\resizebox{\textwidth}{!}{%
\begin{tabular}{l ccccccccc c}
\toprule
\textbf{Model} & \textbf{Retr.} & \textbf{Comm.} & \textbf{Det.} & \textbf{Spec.} & \textbf{Sent.} & \textbf{NetZ.} & \textbf{TCFD} & \textbf{WFB} & \textbf{WXI} & \textbf{Avg.} \\
\midrule
\rowcolor{ctxgray}\multicolumn{11}{l}{\textit{ModernBERT-base}} \\
$\theta_{\textsc{base}}$ & $86.7_{\pm 1.9}$ & $65.9_{\pm 1.3}$ & $94.5_{\pm 0.3}$ & $67.3_{\pm 0.7}$ & $71.8_{\pm 4.1}$ & $98.7_{\pm 0.6}$ & $60.1_{\pm 0.7}$ & $96.0_{\pm 0.4}$ & $20.4_{\pm 5.0}$ & $73.5_{\pm 1.7}$ \\
\midrule
\rowcolor{blue!7}
\multicolumn{11}{l}{\textit{Original RoBERTa model}} \\
ClimateBERT 
& $82.4_{\pm 2.6}$ 
& $70.9_{\pm 0.0}$ 
& $97.5_{\pm 0.0}$ 
& $67.8_{\pm 0.0}$ 
& $76.6_{\pm 0.0}$ 
& $99.1_{\pm 0.0}$ 
& $51.4_{\pm 0.1}$ 
& $92.5_{\pm 0.0}$ 
& $3.3_{\pm 4.6}$ 
& $72.1_{\pm 0.8}$ \\
\midrule

\rowcolor{yellow!10}
\multicolumn{11}{l}{\textit{Climate-ModernBERT: CX + LRD specialization — $\theta_{\mathcal{X}}^{\textsc{lrd}}$}} \\
$\{\mathcal{A}\}$              & $81.1_{\pm 4.1}$ & $68.7_{\pm 2.1}$ & $93.6_{\pm 0.2}$ & $67.3_{\pm 1.1}$ & $77.1_{\pm 0.7}$ & $98.0_{\pm 1.2}$ & $61.3_{\pm 0.5}$ & $96.9_{\pm 0.3}$ & $25.2_{\pm 3.2}$ & $74.4_{\pm 2.4}$ \\
$\{\mathcal{S}\}$              & $84.5_{\pm 2.1}$ & $66.4_{\pm 3.4}$ & $93.8_{\pm 0.2}$ & $69.4_{\pm 0.9}$ & $74.8_{\pm 2.7}$ & $99.0_{\pm 0.1}$ & $61.1_{\pm 0.2}$ & $96.5_{\pm 0.3}$ & $25.0_{\pm 2.4}$ & $74.5_{\pm 1.4}$ \\
$\{\mathcal{F}\}$              & $84.2_{\pm 2.5}$ & $68.0_{\pm 1.6}$ & $95.4_{\pm 0.0}$ & $69.5_{\pm 1.9}$ & $76.9_{\pm 0.4}$ & $98.9_{\pm 0.0}$ & $59.6_{\pm 1.7}$ & $97.1_{\pm 0.2}$ & $20.3_{\pm 4.4}$ & $74.4_{\pm 1.4}$ \\
$\{\mathcal{A},\mathcal{S}\}$  & $83.2_{\pm 4.1}$ & $66.1_{\pm 1.0}$ & $93.2_{\pm 0.8}$ & $69.5_{\pm 1.0}$ & $\mathbf{78.1}_{\pm 0.5}$ & $98.9_{\pm 0.2}$ & $61.3_{\pm 1.0}$ & $96.2_{\pm 0.4}$ & $24.3_{\pm 4.9}$ & $74.5_{\pm 1.5}$ \\
$\{\mathcal{A},\mathcal{F}\}$  & $80.7_{\pm 4.4}$ & $68.8_{\pm 0.6}$ & $94.3_{\pm 0.6}$ & $68.9_{\pm 0.7}$ & $75.1_{\pm 0.4}$ & $99.0_{\pm 0.1}$ & $61.5_{\pm 2.5}$ & $96.5_{\pm 0.5}$ & $\mathbf{27.9}_{\pm 5.0}$ & $74.7_{\pm 1.6}$ \\
$\{\mathcal{S},\mathcal{F}\}$  & $85.3_{\pm 2.2}$ & $62.2_{\pm 2.3}$ & $\mathbf{95.7}_{\pm 0.2}$ & $71.0_{\pm 0.8}$ & $75.5_{\pm 0.6}$ & $97.9_{\pm 0.3}$ & $59.4_{\pm 2.2}$ & $96.9_{\pm 0.5}$ & $23.4_{\pm 3.7}$ & $74.1_{\pm 1.4}$ \\
$\{\mathcal{A},\mathcal{S},\mathcal{F}\}$ & $\mathbf{86.4}_{\pm 1.4}$ & $59.5_{\pm 2.3}$ & $95.5_{\pm 0.0}$ & $\mathbf{71.5}_{\pm 1.4}$ & $75.9_{\pm 1.2}$ & $98.5_{\pm 0.4}$ & $\mathbf{62.4}_{\pm 2.2}$ & $\mathbf{97.3}_{\pm 0.2}$ & $24.4_{\pm 5.9}$ & $74.6_{\pm 1.7}$ \\
\midrule

\rowcolor{yellow!10}\multicolumn{11}{l}{\textit{Ablation: Context Extension-only checkpoints — $\theta_{\mathcal{X}}$}} \\
$\{\mathcal{A}\}$              & $85.5_{\pm 1.0}$ & $\mathbf{71.8}_{\pm 1.1}$ & $93.2_{\pm 0.3}$ & $68.9_{\pm 1.6}$ & $77.4_{\pm 1.0}$ & $98.5_{\pm 0.8}$ & $59.0_{\pm 0.9}$ & $97.0_{\pm 0.2}$ & $26.1_{\pm 3.2}$ & $\mathbf{75.3}_{\pm 1.1}$ \\
$\{\mathcal{S}\}$              & $86.0_{\pm 1.8}$ & $65.2_{\pm 1.0}$ & $93.5_{\pm 0.8}$ & $68.0_{\pm 0.6}$ & $71.8_{\pm 4.1}$ & $98.7_{\pm 0.4}$ & $60.0_{\pm 0.7}$ & $97.2_{\pm 0.2}$ & $20.0_{\pm 2.1}$ & $73.4_{\pm 1.3}$ \\
$\{\mathcal{F}\}$              & $84.5_{\pm 1.8}$ & $67.5_{\pm 0.3}$ & $93.7_{\pm 0.6}$ & $69.5_{\pm 0.8}$ & $76.1_{\pm 0.6}$ & $98.9_{\pm 0.2}$ & $56.4_{\pm 1.0}$ & $96.8_{\pm 0.4}$ & $15.0_{\pm 1.2}$ & $73.2_{\pm 0.8}$ \\
$\{\mathcal{A},\mathcal{S}\}$  & $86.1_{\pm 1.9}$ & $67.9_{\pm 1.4}$ & $94.0_{\pm 0.6}$ & $67.6_{\pm 0.9}$ & $77.0_{\pm 2.0}$ & $98.2_{\pm 0.9}$ & $61.4_{\pm 0.5}$ & $96.6_{\pm 0.5}$ & $24.8_{\pm 3.4}$ & $74.8_{\pm 1.3}$ \\
$\{\mathcal{A},\mathcal{F}\}$  & $86.0_{\pm 0.8}$ & $65.9_{\pm 0.7}$ & $93.8_{\pm 0.4}$ & $67.2_{\pm 1.1}$ & $76.8_{\pm 1.0}$ & $99.1_{\pm 0.2}$ & $59.7_{\pm 1.2}$ & $96.5_{\pm 0.3}$ & $24.1_{\pm 3.2}$ & $74.3_{\pm 1.0}$ \\
$\{\mathcal{S},\mathcal{F}\}$  & $85.4_{\pm 1.0}$ & $64.8_{\pm 3.4}$ & $94.5_{\pm 0.3}$ & $68.0_{\pm 1.5}$ & $76.3_{\pm 0.5}$ & $\mathbf{99.2}_{\pm 0.1}$ & $59.1_{\pm 1.0}$ & $97.2_{\pm 0.1}$ & $23.4_{\pm 5.1}$ & $74.2_{\pm 1.4}$ \\
$\{\mathcal{A},\mathcal{S},\mathcal{F}\}$ & $86.0_{\pm 1.0}$ & $67.1_{\pm 1.3}$ & $95.0_{\pm 0.1}$ & $70.0_{\pm 0.2}$ & $74.9_{\pm 0.6}$ & $99.0_{\pm 0.1}$ & $58.9_{\pm 1.1}$ & $96.3_{\pm 0.9}$ & $18.9_{\pm 5.3}$ & $74.0_{\pm 1.2}$ \\
\bottomrule
\end{tabular}}
\caption{
Per-task F$_1$ scores (\%, mean $\pm$ standard deviation across three random seeds) 
for jointly trained climate-adapted variants. 
LRD-specialized checkpoints represent our primary climate-adapted models following the ModernBERT two-stage training procedure, while context extension-only checkpoints are reported as an ablation to isolate the contribution of LRD specialization.
Binary tasks and ClimRetrieve report positive-class F$_1$; classification tasks report macro-F$_1$. Best results for each task are shown in \textbf{bold}.}
\vspace{-2ex}
\label{tab:data_combinations}
\end{table*}

\paragraph{Evaluation budget.}
In total, we evaluate $21$ climate-adapted checkpoints against $\theta_{\textit{base}}$ on the downstream climate tasks described below. To support the paired-seed ablation analysis of synthetic data, the four checkpoints in the $\{\mathcal{A}\}$/$\{\mathcal{A},\mathcal{S}\}$ comparison pair (with and without LRD specialization) are fine-tuned with $n{=}10$ random seeds, whereas all remaining variants use $n{=}3$ seeds.

\subsection{Downstream Tasks}
\label{sec:tasks}
We evaluate all models on nine climate NLP benchmarks. These include six single-label classification tasks: \textbf{Climate Detection} (Det.) \citep{bingler2024cheap}, \textbf{Climate Specificity} (Spec.) \citep{bingler2024cheap}, \textbf{Commitments \& Actions} (Comm.) \citep{bingler2024cheap}, \textbf{Climate Sentiment} (Sent.) \citep{bingler2024cheap}, \textbf{Net Zero \& Reduction} (NetZ.) \citep{schimanski2023climatebert}, and \textbf{TCFD Recommendations} (TCFD) \citep{bingler2024cheap}. We further evaluate two multi-label classification tasks, \textbf{WFB Nature} (WFB) \citep{schimanski2023exploring} and \textbf{WXImpactBench} (WXI) \citep{yu2025wximpactbench}, as well as one retrieval benchmark, \textbf{ClimRetrieve} (Retr.) \citep{schimanski2024climretrieve}, formulated as binary relevance classification.
Acronyms in parentheses are used throughout the results section (\S\ref{sec:results}). 
Taken together, these benchmarks cover a broad range of climate NLP settings, including topical detection, disclosure analysis, climate commitments, sentiment analysis, information retrieval, and environmental impact assessment. The underlying datasets span corporate annual reports, sustainability disclosures, national policy pledges, and historical news articles, capturing both structured disclosure language and narrative climate communication \citep{calamai2025benchmarking}. Detailed dataset statistics, label definitions, and data provenance are provided in Appendix~\ref{app:tasks}.

\paragraph{Fine-tuning protocol.}
For each ($\theta_{\mathcal{X}}$, task) pair, we jointly fine-tune all encoder parameters together with a task-specific classification head. Following \citet{sounack2025bioclinical}, all variants use a shared hyperparameter configuration: learning rate 4e-5, per-device batch size $32$ with $2$ gradient accumulation steps (effective batch size $64$), weight decay $0.01$, and up to $10$ training epochs with early stopping based on validation F$_1$. Training is performed in BF16 using the fused AdamW optimizer. 
We report test-set performance using F$_1$-based metrics. Binary classification tasks and ClimRetrieve are evaluated using positive-class F$_1$, while multi-class and multi-label tasks are evaluated using macro-F$_1$. The same fine-tuning configuration is applied to ClimateBERT.

\section{Experimental Results}
\label{sec:results}
\S\ref{sec:data_composition} examines how corpus composition affects climate adaptation and reports a CX-only ablation to analyze the contribution of LRD specialization, including a paired-seed ablation of the synthetic corpus. \S\ref{sec:merge} compares parameter-space merging against joint training under the same data budget and uses drop-one merges to attribute per-corpus contributions.

\subsection{Joint Continued Pretraining}
\label{sec:data_composition}

\paragraph{Academic corpora provide the strongest CPT recipe, yet broader mixtures can hurt performance.}
As shown in Table~\ref{tab:data_combinations}, climate adaptation generally improves aggregate performance relative to $\theta_{\textsc{base}}$, despite task-specific trade-offs. Most adapted variants obtain average F$_1$ scores at or above the baseline, with the highest average among the jointly trained variants achieved by CX-only $\theta_{\{\mathcal{A}\}}$ ($75.3$). Improvements are concentrated on register-sensitive tasks, including \textit{Sentiment} (+6.3), \textit{Commitments} (+5.9), \textit{WXImpactBench} (+7.5), and \textit{Specificity} (+4.2), while already saturated benchmarks such as \textit{Detection} (+1.2) and \textit{NetZero} (+0.5) shift only marginally, consistent with observations from \citet{calamai2025benchmarking}.

Notably, adding broader data sources to $\mathcal{A}$ does not improve performance. The average CX-only F$_1$ declines from $75.3$ ($\theta_{\{\mathcal{A}\}}$) to $74.8$ ($\theta_{\{\mathcal{A},\mathcal{S}\}}$), and further drops to $74.0$ on $\theta_{\{\mathcal{A},\mathcal{S},\mathcal{F}\}}$. This effect is particularly pronounced on \textit{Commitments \& Actions}, where adding synthetic and web corpora to the academic training set lowers performance from $71.8$ to $67.1$, suggesting interference from out-of-register sources \citep{kurfali2025climateeval}. The overall pattern further indicates that adaptation benefits are driven by source--task alignment, consistent with observations from domain-adaptive pretraining studies \citep{gururangan2020don}.
Different data combinations excel on different benchmarks.
The CX-only $\theta_{\{\mathcal{A}\}}$ checkpoint achieves the highest score on \textit{Commitments}, while $\theta_{\{\mathcal{A},\mathcal{F}\}}^{\textsc{lrd}}$ achieves the highest score on \textit{WXImpactBench}.
% the web--synthetic combination reaches the top results on \textit{Detection} ($\theta_{\{\mathcal{S},\mathcal{F}\}}^{\textsc{lrd}}$) and \textit{NetZero} ($\theta_{\{\mathcal{S},\mathcal{F}\}}$); $\theta_{\{\mathcal{A},\mathcal{S}\}}^{\textsc{lrd}}$ performs best on \textit{Sentiment}; and $\theta_{\{\mathcal{A},\mathcal{S},\mathcal{F}\}}^{\textsc{lrd}}$ obtains the strongest results on \textit{ClimRetrieve}, \textit{Specificity}, \textit{TCFD}, and \textit{WFB}.

\begin{table}[t]
\centering
\small
\setlength{\tabcolsep}{3.5pt}
\begin{tabular}{l rcr | rcr}
\toprule
& \multicolumn{3}{c|}{\textbf{CX-only ($\theta_{\mathcal{X}}$)}}
& \multicolumn{3}{c}{\textbf{LRD specialization ($\theta_{\mathcal{X}}^{\textsc{lrd}}$)}} \\
\textbf{Task} & $\Delta$ & $p_t$ & $p_w$ & $\Delta$ & $p_t$ & $p_w$ \\
\midrule
Retrieval     & \diffup{0.6} & $.31$ & $.26$
              & \diffup{2.2} & $.66$ & $.77$ \\

Commitments   & \diffdown{3.9} & $\mathbf{<.001}$ & $\mathbf{.002}$
              & \diffdown{2.6} & $\mathbf{.016}$  & $\mathbf{.037}$ \\

Detection     & \diffup{0.8}   & $\mathbf{.012}$  & $\mathbf{.020}$
              & \diffdown{0.4} & $.18$ & $.38$ \\

Specificity   & \diffdown{1.3} & $.07$ & $.13$
              & \diffup{2.2}   & $\mathbf{.001}$  & $\mathbf{.002}$ \\

Sentiment     & \diffdown{0.4} & $.43$ & $.56$
              & \diffup{0.9}   & $\mathbf{.007}$  & $\mathbf{.006}$ \\

NetZero       & \diffdown{0.3} & $.51$ & $.63$
              & \diffup{0.9}   & $.07$ & $.08$ \\

TCFD          & \diffup{2.4}   & $\mathbf{<.001}$ & $\mathbf{.002}$
              & \diffdown{0.0} & $.96$ & $.92$ \\

WFB           & \diffdown{0.4} & $.06$ & $.08$
              & \diffdown{0.8} & $\mathbf{.001}$  & $\mathbf{.004}$ \\

WXImpact      & \diffdown{1.3} & $.25$ & $.16$
              & \diffdown{0.9} & $.49$ & $.63$ \\
\bottomrule
\end{tabular}
\caption{Paired-seed synthetic-data ablation ($n{=}10$). $\Delta$ denotes the mean effect of adding $\mathcal{S}$; $p_t$ and $p_w$ are paired-$t$ and Wilcoxon signed-rank $p$-values. Bold indicates significance ($\alpha{=}0.05$) under both tests.}
\vspace{-2ex}
\label{tab:syn_ablation}
\end{table}

\begin{table*}[t]
\centering
\small
\setlength{\tabcolsep}{4pt}
\renewcommand{\arraystretch}{1.3}
\resizebox{\textwidth}{!}{%
\begin{tabular}{l ccccccccc c}
\toprule
\textbf{Model} & \textbf{Retr.} & \textbf{Comm.} & \textbf{Det.} & \textbf{Spec.} & \textbf{Sent.} & \textbf{NetZ.} & \textbf{TCFD} & \textbf{WFB} & \textbf{WXI} & \textbf{Avg.} \\
\midrule
\rowcolor{ctxgray}\multicolumn{11}{l}{\textit{Joint training on the union — $\theta_{\{\mathcal{A},\mathcal{S},\mathcal{F}\}}^{\textsc{lrd}}$}} \\
$\{\mathcal{A},\mathcal{S},\mathcal{F}\}$
& $\mathbf{86.4}_{\pm 1.4}$
& $59.5_{\pm 2.3}$
& $\mathbf{95.5}_{\pm 0.0}$
& $\mathbf{71.5}_{\pm 1.4}$
& $75.9_{\pm 1.2}$
& $98.5_{\pm 0.4}$
& $62.4_{\pm 2.2}$
& $\mathbf{97.3}_{\pm 0.2}$
& $24.4_{\pm 5.9}$
& $74.6_{\pm 1.7}$ \\
\midrule
\rowcolor{ctxgray}\multicolumn{11}{l}{\textit{Parameter-space merging of the three single-source components}} \\
$\theta_{\textsc{Soup}}$        & $83.6_{\pm 1.6}$ & $66.5_{\pm 3.7}$ & $95.4_{\pm 0.6}$ & $70.9_{\pm 2.1}$ & $76.7_{\pm 1.4}$ & $98.9_{\pm 0.0}$ & $\mathbf{64.1}_{\pm 2.2}$ & $97.3_{\pm 0.1}$ & $\mathbf{33.7}_{\pm 5.3}$ & $\mathbf{76.3}_{\pm 1.9}$ \\
$\theta_{\textsc{TA}(0.5)}$     & $80.9_{\pm 5.7}$ & $67.5_{\pm 1.8}$ & $94.7_{\pm 0.4}$ & $67.9_{\pm 0.2}$ & $71.0_{\pm 0.2}$ & $99.1_{\pm 0.0}$ & $60.3_{\pm 2.4}$ & $95.8_{\pm 0.5}$ & $24.8_{\pm 6.0}$ & $73.6_{\pm 1.9}$ \\
$\theta_{\textsc{TA}(1.0)}$     & $85.0_{\pm 0.4}$ & $71.2_{\pm 1.2}$ & $94.9_{\pm 0.5}$ & $68.5_{\pm 0.7}$ & $77.0_{\pm 3.1}$ & $\mathbf{99.2}_{\pm 0.1}$ & $61.3_{\pm 0.7}$ & $97.1_{\pm 0.2}$ & $27.2_{\pm 5.1}$ & $75.7_{\pm 1.3}$ \\
$\theta_{\textsc{TIES}(0.5)}$   & $86.1_{\pm 1.4}$ & $67.6_{\pm 2.5}$ & $93.6_{\pm 0.6}$ & $68.9_{\pm 0.9}$ & $\mathbf{77.1}_{\pm 0.7}$ & $99.0_{\pm 0.1}$ & $61.4_{\pm 0.6}$ & $96.5_{\pm 1.2}$ & $28.5_{\pm 1.7}$ & $75.4_{\pm 1.1}$ \\
$\theta_{\textsc{TIES}(0.7)}$   & $85.1_{\pm 2.8}$ & $\mathbf{71.9}_{\pm 2.7}$ & $94.9_{\pm 0.1}$ & $70.7_{\pm 0.1}$ & $75.6_{\pm 1.6}$ & $99.0_{\pm 0.2}$ & $59.7_{\pm 1.1}$ & $96.7_{\pm 1.3}$ & $26.9_{\pm 3.8}$ & $75.6_{\pm 1.5}$ \\
$\theta_{\textsc{DARE}(0.5)}$   & $86.3_{\pm 0.9}$ & $67.2_{\pm 0.4}$ & $94.8_{\pm 0.3}$ & $69.2_{\pm 0.2}$ & $74.3_{\pm 0.1}$ & $98.9_{\pm 0.2}$ & $57.8_{\pm 1.3}$ & $97.1_{\pm 0.2}$ & $26.9_{\pm 2.3}$ & $74.7_{\pm 0.7}$ \\
$\theta_{\textsc{DARE}(0.7)}$   & $85.5_{\pm 0.9}$ & $63.3_{\pm 1.1}$ & $94.7_{\pm 0.1}$ & $67.6_{\pm 0.3}$ & $74.1_{\pm 2.0}$ & $98.9_{\pm 0.2}$ & $61.6_{\pm 1.0}$ & $97.0_{\pm 0.3}$ & $26.2_{\pm 6.2}$ & $74.3_{\pm 1.3}$ \\
% $\theta_{\textsc{A2x}}$         & $\mathbf{87.0}$ & $67.6$          & $94.3$          & $67.3$          & $75.2$          & $99.0$          & $60.1$          & $96.7$          & $21.6$          & $74.9$ \\
% % $\theta_{\textsc{A3x}}$         & $85.6$          & $68.0$          & $94.1$          & $66.8$          & $77.0$          & $97.9$          & $60.2$          & $\mathbf{97.1}$ & $22.3$          & $74.6$ \\
% $\theta_{\textsc{Fhalf}}$       & $86.4$          & $67.1$          & $94.7$          & $\mathbf{68.3}$ & $74.2$          & $99.0$          & $58.7$          & $96.4$          & $\mathbf{28.9}$ & $\mathbf{75.6}$ \\
% $\theta_{\textsc{NormBal}}$     & $84.3$          & $67.3$          & $94.1$          & $67.5$          & $\mathbf{78.0}$ & $99.0$          & $\mathbf{61.4}$ & $96.4$          & $23.8$          & $75.2$ \\
% \midrule
% \rowcolor{ctxgray}\multicolumn{11}{l}{\textit{Parameter-space merging of CX-only single-source components}} \\
% $\theta_{\textsc{SOUP}}$                    & $84.5$          & $66.5$          & $94.1$          & $\mathbf{69.6}$ & $78.2$          & $\mathbf{98.3}$ & $\mathbf{61.3}$ & $96.5$          & $22.5$          & $74.6$ \\
% $\theta_{\textsc{Fhalf}}$     & $85.1$          & $\mathbf{70.5}$ & $92.9$          & $68.2$          & $\mathbf{78.6}$ & $98.1$          & $61.0$          & $96.3$          & $20.9$          & $74.6$ \\
% $\theta_{\textsc{NormBal}}$   & $\mathbf{86.5}$ & $70.1$          & $\mathbf{94.4}$ & $68.8$          & $77.8$          & $\mathbf{98.3}$ & $61.2$          & $\mathbf{96.6}$ & $\mathbf{26.8}$ & $\mathbf{75.6}$ \\
\bottomrule
\end{tabular}}
\caption{Joint training on the full corpus union and seven parameter-space merges of the corresponding single-source LRD checkpoints, reporting per-task F1 (\%, mean ± standard deviation across three random seeds).}
\label{tab:merging}
\vspace{-3ex}
\end{table*}

\paragraph{Synthetic climate text exhibits task-dependent effects.}
To isolate the effect of synthetic data, we compare $\theta_{{\mathcal{A}}}$ and $\theta_{{\mathcal{A},\mathcal{S}}}$ under both the final LRD-specialized models and the CX-only ablation setting using $n{=}10$ shared fine-tuning seeds. As shown in Table~\ref{tab:syn_ablation}, the effect of $\mathcal{S}$ is strongly task dependent. Adding synthetic data significantly improves several framework-oriented benchmarks, including \textit{TCFD} ($+2.4$, $p_t<.001$) and \textit{Specificity} ($+2.2$, $p_t=.001$), while consistently degrading \textit{Commitments \& Actions} across both training regimes ($-3.9$, $p_t<.001$). In contrast, retrieval performance remains statistically unchanged. Overall, these results suggest that synthetic climate text benefits taxonomy- and framework-driven classification tasks, but transfers less effectively to tasks requiring finer-grained discourse and commitment understanding.

\begin{figure}[t]
    \centering
    \includegraphics[width=0.9\columnwidth]{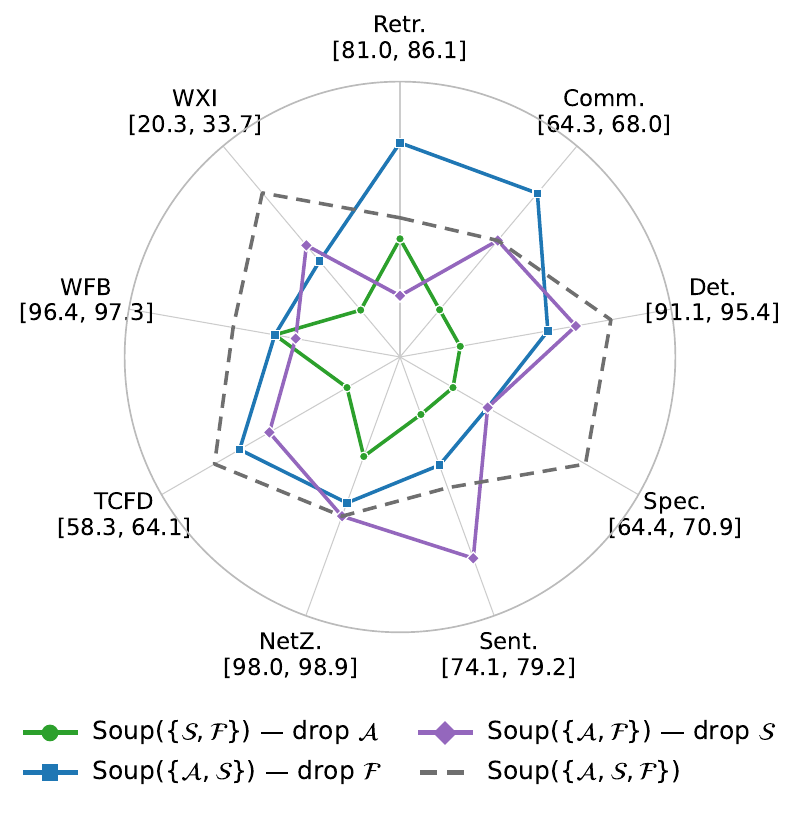}
    \caption{Per-task F$_1$ of the three drop-one Soups against the full Soup($\{\mathcal{A},\mathcal{S},\mathcal{F}\}$) baseline (dashed grey). 
    % Each coloured polygon corresponds to the merge that \emph{excludes} one corpus: green = academic dropped, blue = FineWeb dropped, purple = synthetic dropped. 
    Axes are scaled per task to the $[\min,\max]$ range across the four variants.}
    \label{fig:drop_one_radar}
    \vspace{-2ex}
\end{figure}

\subsection{Parameter-space Merging}
\label{sec:merge}

We evaluate model merging both as a data-attribution method and as an alternative to joint training under matched data coverage. Table~\ref{tab:merging} compares joint training on the full corpus
union with seven parameter-space merging variants, while Figure~\ref{fig:drop_one_radar} presents drop-one Soup ablations relative to the full $\theta_{\textsc{Soup}}(\{\mathcal{A},\mathcal{S},\mathcal{F}\})$ model. We additionally evaluate model merging using CX-only checkpoints ($\theta_{\mathcal{X}}$). These results are reported in Appendix~\ref{app:cx-merging}.

\paragraph{Dropping $\mathcal{A}$ hurts more than dropping either other corpus.}

Figure~\ref{fig:drop_one_radar} provides a complementary perspective on corpus importance through drop-one simple weight averaging ablations. Across nearly all benchmarks, removing the academic corpus (\textsc{Soup}($\{\mathcal{S},\mathcal{F}\}$), green) results in the largest performance degradation, reducing average F$_1$ by $4.0$ points relative to the full \textsc{Soup}($\{\mathcal{A},\mathcal{S},\mathcal{F}\}$). In contrast, excluding either the synthetic or web corpus leads to substantially smaller average degradations of $1.5$ points, and in some cases even improves performance on individual benchmarks. These results provide independent evidence for the findings in \S\ref{sec:data_composition}: academic climate data contributes the strongest and most broadly transferable adaptation signal, whereas the benefits of synthetic and web data are comparatively task-dependent.

Interestingly, this importance is not reflected by task-vector magnitude alone. Although the FineWeb-specialized checkpoint produces the largest parameter update ($\|\boldsymbol{\delta}_{F}\| \approx 43$ per layer, compared with approximately $16$ for $\mathcal{A}$ and $\mathcal{S}$), removing $\mathcal{A}$ causes by far the largest performance degradation. This observation indicates that larger parameter updates do not necessarily translate into greater downstream utility, highlighting an interesting direction for future work on interpreting adaptation dynamics in continued pretraining.

\begin{figure}[t]
\centering
\includegraphics[width=0.85\columnwidth]{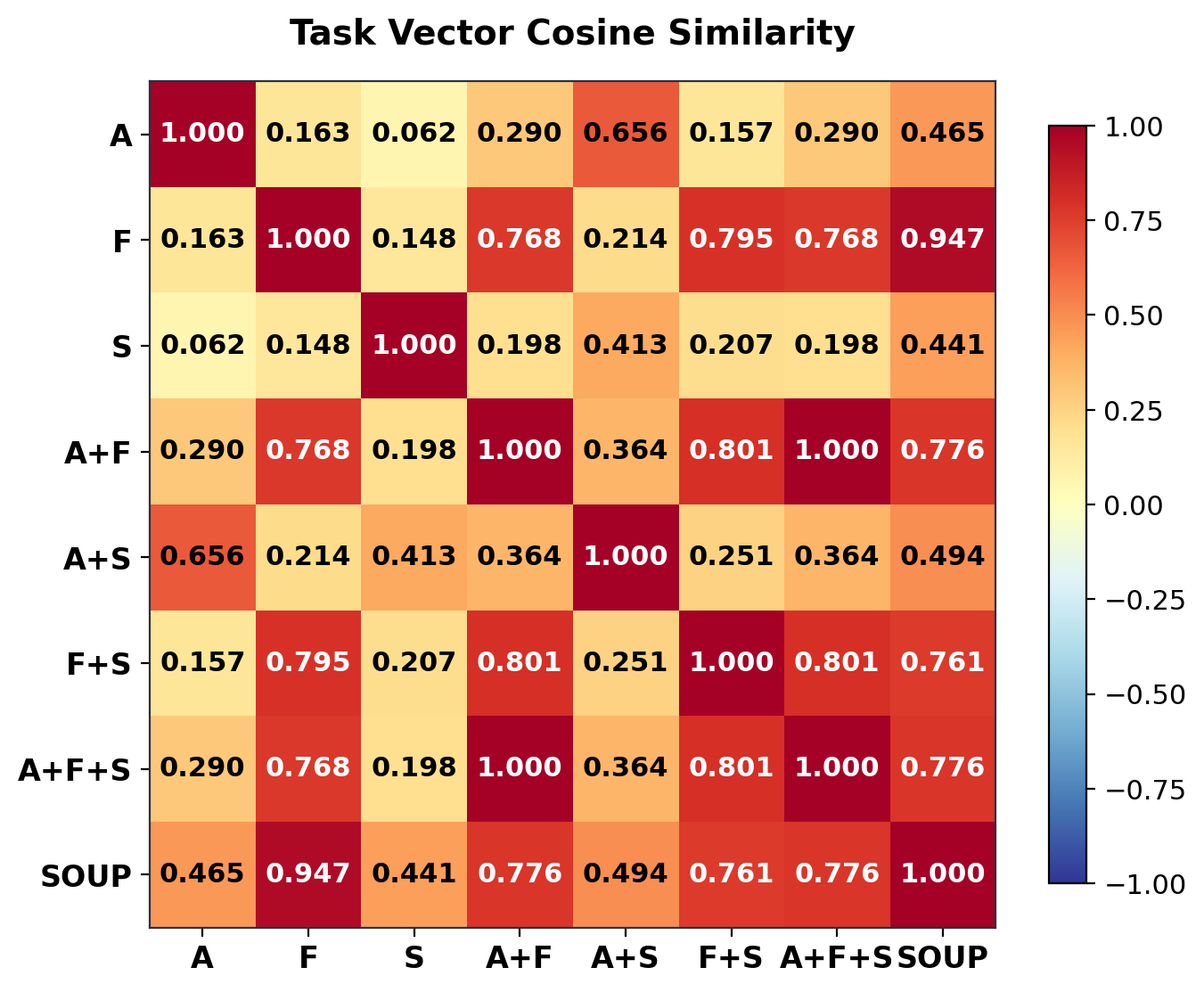}
\caption{Cosine similarity between source task vectors. 
}
\label{fig:cosine_sim}
\vspace{-3ex}

\end{figure}

\paragraph{Merging surpasses both the strongest single source and joint training on the same data.}
Across the seven merged models in Table~\ref{tab:merging}, five outperform the strongest single-source LRD checkpoint ($\theta_{\{\mathcal{S}\}}^{\textsc{lrd}}$, $74.5$), with uniform weight averaging performing best overall. $\theta_{\textsc{Soup}}$ achieves $76.3$ average F$_1$, improving by $1.8$ points over the best individual source and by $1.7$ points over joint training on the same effective corpus. The gains are most pronounced on \textit{Commitments} ($+7.0$) and \textit{WXImpactBench} ($+9.3$), while joint training retains only narrow advantages on \textit{Detection}, \textit{Specificity}, and \textit{Retrieval}. Although weight averaging is known to improve performance in supervised fine-tuning settings \citep{wortsman2022model}, evidence in the context of multi-source continued pretraining remains limited.

To better understand this behavior, we analyze the task vectors $\boldsymbol{\delta}_i = \theta_{\textsc{lrd},i} - \theta_{\textit{base}}$ of the three source-specific checkpoints. Figure~\ref{fig:cosine_sim} shows that the resulting parameter updates are nearly orthogonal, with pairwise cosine similarities ranging from $0.06$ to $0.18$ between $\mathcal{A}$, $\mathcal{F}$, and $\mathcal{S}$. Such weak alignment indicates that each corpus contributes largely distinct parameter updates, consistent with observations from task-vector analyses in fine-tuning \citep{ilharco2023editing}. The complementarity of these updates provides a plausible explanation for why parameter-space merging can recover information from multiple sources more effectively than joint optimization.

% To understand the structure underlying these merges, we analyze the task vectors
% $\boldsymbol{\delta}_i = \theta_{\text{LRD}_i} - \theta_{\text{BASE}}$ of each
% source-specific model. As shown in Figure~\ref{fig:cosine_sim}, the three core
% sources have small pairwise cosine similarities in weight space (0.06–0.18 between $\mathcal{A}$
% , $\mathcal{F}$
% , $\mathcal{S}$
% ), consistent with the close-orthogonality of task vectors reported in other fine-tuning settings \citep{ilharco2023editing}.

% Despite this complementarity, the sources differ sharply in magnitude: F's task
% vector ($\lVert\boldsymbol{\delta}_F\rVert \approx 43$ per layer) is
% 2.5--3$\times$ larger than A's or S's ($\lVert\boldsymbol{\delta}\rVert \approx 16$).
% In a uniform average, F therefore dominates the merged model (cosine similarity
% between $\delta_{\text{SOUP}}$ and $\delta_F$ is 0.85), despite the
% leave-one-out analysis showing that A's removal causes the largest performance
% degradation ($-4.0$ F$_1$). 

\section{Conclusion}
We introduce \textsc{Climate-ModernBERT}, a collection of climate-adapted encoder models obtained through ModernBERT's continued pretraining procedure with LRD specialization of ModernBERT-Base on academic climate corpora, climate-filtered FineWeb-Edu web data, and synthetic climate text.
Across 21 adapted variants and nine climate NLP benchmarks, we reveal three main findings: 
\textit{(1)} academic climate data provides the strongest adaptation signal, while broader corpus mixtures often fail to improve and can degrade performance; 
\textit{(2)} synthetic climate text exhibits task-dependent effects, benefiting taxonomy-driven tasks yet hurting performance on pragmatically grounded benchmarks; 
and \textit{(3)} simple weight averaging of independently specialized checkpoints achieves the best overall performance (76.3 average F$_1$), outperforming both single-source models and joint training on the union of the full corpora.

These results highlight the importance of corpus composition in domain-adaptive pretraining and suggest that parameter-space merging is a simple and effective alternative to joint multi-source training. 
We release all \textsc{Climate-ModernBERT} variants, training checkpoints, and data-processing pipelines to support future research in climate NLP and domain-adaptive encoder pretraining.

\section*{Limitations}
Our study has several limitations. To the best of our knowledge and recent literature \citep{calamai2025benchmarking}, current climate NLP benchmarks remain primarily designed around sentence- or passage-level classification tasks over corporate disclosures, sustainability reports, and policy documents. Although ModernBERT is designed for long-context processing, existing benchmark suites provide limited opportunities to evaluate long-document understanding, cross-section reasoning, or information integration across extended contexts beyond ClimRetrieve. Consequently, the potential benefits of long-context architectures cannot yet be fully assessed within the current climate NLP evaluation landscape. We therefore view the development of standardized long-context and document-centric climate NLP benchmarks as an important direction for future research.

Furthermore, \textsc{Climate-ModernBERT} collection is trained exclusively on English climate text and adapted from a single encoder family, ModernBERT-Base. Our analysis focuses on climate NLP and evaluates adaptation strategies using climate-specific corpora and benchmarks. Therefore, the observed effects of corpus composition and parameter-space merging should be interpreted as findings within this domain rather than universal principles of domain adaptation. Extending this analysis to other specialized domains, such as biomedical or legal NLP, remains an important direction for future work. More broadly, our analysis is restricted to continued pretraining. We do not examine instruction-tuned encoder variants \citep{clavie2025all}, whose representations and adaptation dynamics may differ substantially from base models. Understanding how corpus composition and parameter-space merging interact with instruction tuning is thus an important direction for future work.

\section*{Ethical Considerations}
Our academic climate corpus is compiled from climate-related scientific publications, reports, and other copyrighted sources. As climate NLP increasingly supports decision-making and policy analysis, we respect licensing and intellectual property restrictions by releasing only trained model checkpoints and data-processing pipelines where permitted, rather than the raw text data.

Climate change is a societal challenge that requires collaboration among governments, industry, researchers, and citizens \citep{reid2009responding}. We hope that climate-adapted language models can support information retrieval, document analysis, and evidence synthesis across large climate-related document collections.

\section*{Acknowledgment}
Financial support from the Swiss National Science Foundation (Grant Agreement No. 207800) is gratefully acknowledged. We also would like to thank Prof. Mrinmaya Sachan from ETH Zurich for his support of computational resources used in this work. We also thank the anonymous ARR reviewers for their constructive comments, which helped improve this manuscript.

\bibliography{custom}

\appendix
\newpage

\section{Data Preprocessing Pipeline}
\label{app:preprocessing}

Both corpora ($\mathcal{A}$, $\mathcal{S}$) pass through a shared deduplication and decontamination pipeline implemented with NeMo Curator \citep{nemo_curator}. Raw documents are read from JSONL and normalized: Unicode characters are reformatted to a canonical encoding, repeated whitespace is collapsed while paragraph breaks are preserved, and URLs are stripped. The climate-filtered web data ($\mathcal{F}$) is preprocessed with MinHash--LSH \citep{rao2016searching} configured with $24$-character $n$-grams, $20$ hash bands, and $13$ MinHash signatures per band, retaining pairs that share approximately $80\%$ of their tokens at a tractable computational cost.

\section{Academic Climate Data: Sources and Statistics}
\label{app:academic-data}

The academic climate corpus combines four streams of in-domain text: peer-reviewed journal articles, year-aggregated climate news, climate-focused arXiv preprints, and a small set of climate handbooks. Table~\ref{tab:academic-sources} reports a category-level summary with the format, number of source titles, approximate document count, and raw token contribution (before deduplication) from each stream.

\begin{table*}[t]
\centering
\small
\setlength{\tabcolsep}{5pt}
\renewcommand{\arraystretch}{1.15}
\begin{tabularx}{\textwidth}{>{\RaggedRight\arraybackslash}X l r r r}
\toprule
\textbf{Source / Category} & \textbf{Fmt} & \textbf{Srcs} & \textbf{Docs} & \textbf{Raw} \\
\midrule
\rowcolor{lightgray}\multicolumn{5}{l}{\textit{Peer-reviewed journals}} \\
Climate / Earth-science journals
  & XML & $\sim$25 titles & $\sim$30K & $\sim$1.70B \\
\textit{e.g., Climate Risk Mgmt, Urban Climate, Env.\ Sci.\ \& Policy, Weather \& Climate Extremes, Atmospheric Env., J.\ of Climate, Sci.\ Total Env., Earth-Science Reviews, Global Env.\ Change.}
  & & & & \\
Energy / resource / env.\ economics
  & XML & $\sim$12 titles & $\sim$15K & $\sim$0.80B \\
\textit{e.g., Energy Economics, Energy Policy, Ecological Economics, Resources Policy, Resource \& Energy Econ., Renewable \& Sustainable Energy Reviews.}
  & & & & \\
Finance / macro / accounting (climate-relevant)
  & XML & $\sim$45 titles & $\sim$40K & $\sim$1.40B \\
\textit{e.g., J.\ Banking \& Finance, J.\ Financial Economics, J.\ Public Econ., J.\ Corporate Finance, Accounting Org.\ \& Society.}
  & & & & \\
\midrule
\rowcolor{lightgray}\multicolumn{5}{l}{\textit{Other in-domain sources}} \\
ClimateNews archive (2000--2022)
  & CSV & 23 shards & $\sim$5M articles & $\sim$0.60B \\
Climate arXiv preprints (physics, econ, cs)
  & PDF$\rightarrow$text & --- & $>$1K & $\sim$0.12B \\
Climate handbooks
  & PDF$\rightarrow$text & 7 vols & 7 books & $\sim$0.025B \\
\midrule
\textbf{Total}
  & --- & --- & --- & $\sim$4.65B \\
\bottomrule
\end{tabularx}
\caption{Per-category composition of the academic climate corpus $\mathcal{A}$. \texttt{Srcs}: number of unique source titles; \texttt{Docs}: number of source documents; \texttt{Raw}: token count before deduplication.}
\label{tab:academic-sources}
\end{table*}

\paragraph{Acquisition and preprocessing.}
We parse journal articles in structured XML directly, retain section labels and captions, and exclude reference lists, author affiliations, and copyright boilerplate. The \textit{ClimateNews} archive is distributed as twenty-three annual CSV shards (one per year, 2000--2022), each of which we filter for climate relevance using the keyword list of Appendix~\ref{app:web-keywords} before tokenization. arXiv preprints are downloaded as PDF and converted to text with a layout-aware extractor that preserves section hierarchy and detects figure/table captions; we restrict the pool to climate-relevant categories (\texttt{physics.ao-ph}, \texttt{econ.GN}, \texttt{q-fin.RM}, and climate-tagged \texttt{cs} papers). The climate handbooks are processed with the same PDF pipeline. Across all four streams we apply boilerplate stripping, English-only language identification, and a $200$-token minimum-length filter before documents enter the deduplication pipeline.

\begin{table}[t]
\centering
\small
\setlength{\tabcolsep}{6pt}
\renewcommand{\arraystretch}{1.15}
\begin{tabular}{l r}
\toprule
\textbf{Category} & \textbf{Seeds} \\
\midrule
Climate Risk Management                 & $654$ \\
Environmental Science                   & $3{,}077$ \\
Renewable \& Sustainable Energy Reviews & $1{,}028$ \\
Urban Climate                           & $541$ \\
Weather and Climate Extremes            & $685$ \\
ClimateNews (2000, 2005, 2017, 2022)    & $4 \times 200$ \\
\midrule
\textbf{Total}                          & $\mathbf{6{,}785}$ \\
\bottomrule
\end{tabular}
\caption{Seed pool used to condition synthetic generation. Each seed yields three outputs (one per style); the resulting raw synthetic corpus before deduplication contains approximately $20$K passages.}
\label{tab:syn-seed-pool}
\end{table}

\paragraph{Licensing and ethics.}
Peer-reviewed articles are accessed under publisher licenses held by our institution; arXiv preprints are released by their authors under permissive arXiv terms; the news shards and climate handbooks were collected and processed for non-commercial research use. The corpus is used solely for research on domain-adaptive pretraining and is not redistributed; only model checkpoints, the keyword list, and aggregate per-source statistics are released. The corpus contains only text already in the public or licensed scientific record and does not specifically target personally identifiable information.

\section{Web Data: Keyword List for FineWeb-Edu Filtering}
\label{app:web-keywords}

The web-data pipeline retains a FineWeb-Edu document if any of the case-insensitive terms below appears in the document (for full list please refer to our github repository). The terms are extracted from the climate concept in General Multilingual Environment Thesaurus (GEMET)\footnote{\url{https://www.eionet.europa.eu/gemet/en/themes/}} and \citet{sautner2023firm}. For deduplication we used the NeMo Curator \citep{nemo_curator} deduplication pipeline.
% and grouped into \textit{strong} and \textit{weak} climate signals. 
% The Stage-2 FastText\footnote{\url{https://github.com/facebookresearch/fastText/}} classifier (\S\ref{sec:data}) is trained on a $10$K subset of the Stage-1 pool annotated for climate relevance with GPT-4o, and applied with a decision threshold of $0.5$ on the predicted climate probability.

\paragraph{Keywords}
{\small\texttt{climate change, global warming, greenhouse gas, GHG, carbon dioxide, CO\textsubscript{2}, methane, nitrous oxide, carbon budget, net zero, decarbonization, renewable energy, carbon capture, carbon sequestration, negative emissions, emissions trading, carbon pricing, carbon neutrality, integrated assessment model, climate scenario, climate modelling, IPCC, El Ni\~no, ENSO, radiative forcing, radiative imbalance, arctic amplification, ocean heat content, sea level rise, permafrost, cryosphere, monsoon, flood risk, storm surge, extreme weather, heatwave, drought, climate hazard, loss and damage, loss and damage fund, 1.5\textdegree C, 2\textdegree C, Paris Agreement, global stocktake, nationally determined contribution, NDC, just transition, transition pathway, climate litigation, climate governance, climate finance, climate justice.}}
\begin{table}[t]
\centering
\small
\setlength{\tabcolsep}{6pt}
\renewcommand{\arraystretch}{1.15}
\begin{tabular}{l c c}
\toprule
\textbf{Task} & $\theta_{\textsc{base}}$ \textbf{(stable-phase)} & \textbf{ModernBERT (post-LRD)} \\
\midrule
Retr.  & $86.7_{\pm 1.9}$ & $83.1_{\pm 1.5}$ \\
Comm.  & $65.9_{\pm 1.3}$ & $60.6_{\pm 1.5}$ \\
Det.   & $94.5_{\pm 0.3}$ & $94.4_{\pm 0.4}$ \\
Spec.  & $67.3_{\pm 0.7}$ & $67.4_{\pm 2.0}$ \\
Sent.  & $71.8_{\pm 4.1}$ & $74.9_{\pm 2.5}$ \\
NetZ.  & $98.7_{\pm 0.6}$ & $98.9_{\pm 0.3}$ \\
TCFD   & $60.1_{\pm 0.7}$ & $57.9_{\pm 0.3}$ \\
WFB    & $96.0_{\pm 0.4}$ & $96.8_{\pm 0.1}$ \\
WXI    & $20.4_{\pm 5.0}$ & $20.2_{\pm 0.7}$ \\
\midrule
\textbf{Avg.}   & $73.5_{\pm 1.7}$ & $72.7_{\pm 1.0}$ \\
\bottomrule
\end{tabular}
\caption{Per-task F$_1$ comparison between the pre-LRD stable-phase ModernBERT-base ($\theta_{\textsc{base}}$) used as initialization throughout the paper and the final publicly released (post-LRD) ModernBERT-base checkpoint. Both models are fine-tuned on the nine climate benchmarks under the same protocol (mean $\pm$ standard deviation across three random seeds).}
\label{tab:modernbert-full}
\end{table}

% \paragraph{Signal-2.}
% {\small\texttt{adaptation, climate resilience, climate risk, climate policy, climate mainstreaming, climate diplomacy, climate accountability, health impact, food security, water security, vector-borne disease, climate migration, disaster risk reduction, compound event, climate system, climate variability, climate dynamics, energy balance, water vapour feedback, atmospheric circulation, paleoclimate, proxy record, temperature rise.}}

\begin{table*}[t]
\centering
\small
\setlength{\tabcolsep}{8pt}
\renewcommand{\arraystretch}{1.15}
\begin{tabular}{ll|ll}
\toprule
\multicolumn{2}{c|}{\textbf{Training Configuration}} &
\multicolumn{2}{c}{\textbf{Model Architecture}} \\
\midrule

\rowcolor{lightgray}
\multicolumn{2}{l|}{\textbf{Context Extension}} &
\multicolumn{2}{l}{\textbf{ModernBERT-Base}} \\

Data & Full climate corpus (\S\ref{sec:data}) &
Parameters & 150M \\

Epochs & 3 &
Layers & 22 \\

Learning rate & $3\times10^{-4}$ (constant) &
Hidden size & 768 \\

Warm-up & 0 epochs &
Intermediate size & 1,152 \\

Global batch size & 576 &
Attention heads & 12 (head dim 64) \\

Per-GPU batch size & 72 &
Vocabulary size & 50,368 \\

Microbatch size & 12 &
Position encoding & RoPE (base 160K) \\

Sequence length & 8,192 &
Attention pattern & Alternating local/global \\

MLM masking rate & 30\% &
Sliding window & 128 \\

Optimizer & StableAdamW &
Global attention & Every 3 layers \\

$\beta_1,\beta_2$ & 0.9, 0.98 &
Optimizations & FlashAttention, unpadding, \texttt{torch.compile} \\

$\epsilon$ & $10^{-6}$ &
& \\

Weight decay & $10^{-5}$ &
\multicolumn{2}{l}{\textbf{Compute \& Checkpointing}} \\

Precision & BF16 (AMP) &
Hardware & $4\times$ NVIDIA A100 \\

\rowcolor{lightgray}
\multicolumn{2}{l|}{\textbf{LRD Specialization}} &
Framework & MosaicML Composer \\

Data & Variant-specific subset &
Mixed precision & BF16 \\

Epochs & 3 &
Gradient checkpointing & Enabled \\

LR schedule & $1-\sqrt{t}$ decay &
Checkpoint (Context Extension) & Every 3,000 batches \\

Initial LR & $3\times10^{-4}$ &
Checkpoint (LRD Specialization) & Every 1,000 batches \\

Final LR factor ($\alpha_f$) & 0.001 &
Release format & HF Transformers FlexBERT \\

Other settings & Same as context extension phase &
& \\

\bottomrule
\end{tabular}
\caption{Training configuration, architecture, and compute details for all \textsc{Climate-ModernBERT} variants.}
\label{tab:full-config}
\end{table*}

\begin{table*}[t]
\centering
\small
\setlength{\tabcolsep}{4pt}
\renewcommand{\arraystretch}{1.15}
\begin{tabular}{@{}>{\raggedright\arraybackslash}p{3.5cm} l c r r >{\raggedright\arraybackslash}p{7.5cm}@{}}
\toprule
\textbf{Task} & \textbf{Type} & \textbf{\#C} & \textbf{\#Train} & \textbf{\#Test} & \textbf{Labels} \\
\midrule
\rowcolor{lightgray}\multicolumn{6}{@{}l}{\textit{Single-label classification}} \\[2pt]
Climate Detection \citep{bingler2024cheap}              & Binary       & 2 & 1{,}170 & 400 & \textit{climate}/\textit{not}: passage is related to climate change or environmental topics. \\
Climate Specificity \citep{bingler2024cheap}            & Binary       & 2 & 900     & 320 & \textit{specific}/\textit{non-specific}: contains concrete, firm-specific commitments or measurable targets vs.\ vague phrasing. \\
Commitments \& Actions \citep{bingler2024cheap}         & Binary       & 2 & 900     & 320 & \textit{yes}/\textit{no}: states a forward-looking pledge or describes a realized climate action. \\
Climate Sentiment \citep{bingler2024cheap}              & Multi-class  & 3 & 900     & 320 & \textit{risk}/\textit{neutral}/\textit{opportunity}: evaluative stance toward climate change in corporate or policy discourse. \\
Net Zero \& Reduction \citep{schimanski2023climatebert} & Multi-class  & 3 & 2{,}753 & 344 & \textit{net-zero}/\textit{reduction}/\textit{none}: paragraph contains a net-zero target, a reduction target, or no target. \\
TCFD Recommendations \citep{bingler2024cheap}           & Multi-class  & 4 & 1{,}170 & 400 & \textit{governance}/\textit{strategy}/\textit{risk}/\textit{metrics-and-targets}: TCFD disclosure pillar. \\
\midrule
\rowcolor{lightgray}\multicolumn{6}{@{}l}{\textit{Multi-label classification}} \\[2pt]
WFB Nature \citep{schimanski2023exploring}              & Multi-label  & 4 & 1{,}760 & 220 & \textit{water}/\textit{forest}/\textit{biodiversity}/\textit{nature}: TNFD-aligned nature dimensions co-occurring in the passage. \\
WXImpactBench \citep{yu2025wximpactbench}               & Multi-label  & 6 & 970     & 208 & \textit{infrastructural}/\textit{political}/\textit{financial}/\textit{ecological}/\textit{agricultural}/\textit{human-health}: societal impact categories of weather events. \\
\midrule
\rowcolor{lightgray}\multicolumn{6}{@{}l}{\textit{Retrieval (binary relevance)}} \\[2pt]
ClimRetrieve \citep{schimanski2024climretrieve}         & Retrieval    & 2 & 6{,}800 & 850 & \textit{relevant}/\textit{not}: passage answers the climate question; relevance binarized at threshold $\geq 1$ on the original ordinal scale. \\
\bottomrule
\end{tabular}
\caption{Overview of downstream evaluation tasks. \textit{\#C} denotes the number of target classes or labels; \textit{\#Train} and \textit{\#Test} report fine-tuning split sizes; the \textit{Labels} column lists the label set followed by a short definition.}
\label{tab:tasks}
\end{table*}
\section{Synthetic Climate Data: Seed Pool, Prompts, and Generation}
\label{app:synthetic-prompts}

Table~\ref{tab:syn-seed-pool} reports the seed pool used for synthetic generation. The five academic seed pools are full subsets of the corresponding journals in $\mathcal{A}$ (\S\ref{sec:data}); the \textit{ClimateNews} pool is sampled at $200$ documents per annual shard. For each seed document, we extract the first $800$ characters of cleaned body text and substitute it into one of three style templates below. Each (seed, style) pair is sent to \texttt{Qwen3.5-122B-A10B} once; the generator is decoded with temperature $0.6$, top-$p$ $0.95$, and a budget of $1{,}024$ new tokens. All generations share the same system prompt: \textit{``You are an expert climate researcher and journalist capable of generating high-quality, factual content about climate change based on real research excerpts.''} The \texttt{ClimateNews} category reuses the same three templates with the lead-in reworded to ``climate news excerpt'' / ``climate news piece''.

\paragraph{Style 1 --- Public awareness.}
\begin{quote}\small\itshape
Based on this climate research excerpt: ``\textless EXTRACT\textgreater''.\\
Create an accessible article about climate change that: (i) \textbf{Explains Complex Concepts}---break scientific terms into everyday language; (ii) \textbf{Local Relevance}---connect global climate issues to local community impacts; (iii) \textbf{Practical Actions}---suggest concrete steps individuals can take; (iv) \textbf{Human Stories}---include relatable examples and potential human impacts. Write in an engaging, informative tone suitable for general public awareness; focus on making climate science understandable and actionable.
\end{quote}

\paragraph{Style 2 --- Industry perspective.}
\begin{quote}\small\itshape
Using this climate-related research as context: ``\textless EXTRACT\textgreater''.\\
Develop a comprehensive industry analysis covering: (i) \textbf{Sector Impact}---how different industries are affected by or contributing to climate change; (ii) \textbf{Innovation \& Technology}---emerging solutions and technological adaptations; (iii) \textbf{Business Strategy}---corporate responses and sustainable business models; (iv) \textbf{Investment Trends}---green finance and ESG considerations. Write from a business and industry perspective, highlighting opportunities and challenges.
\end{quote}

\paragraph{Style 3 --- Environmental journalism.}
\begin{quote}\small\itshape
Drawing from this climate research piece: ``\textless EXTRACT\textgreater''.\\
Create an in-depth environmental journalism article that: (i) \textbf{Investigative Depth}---explore underlying causes and systemic issues; (ii) \textbf{Ecosystem Impact}---detail effects on biodiversity, habitats, and natural systems; (iii) \textbf{Climate Justice}---address equity and vulnerable-population concerns; (iv) \textbf{Solution Spotlight}---highlight successful interventions and best practices. Write in an investigative journalism style that combines factual reporting with compelling narrative.
\end{quote}

\begin{table*}[t]
\centering
\small
\renewcommand{\arraystretch}{1.25}
\resizebox{\textwidth}{!}{%
\begin{tabular}{l ccccccccc c}
\toprule
\textbf{Model} 
& \textbf{Retr.} 
& \textbf{Comm.} 
& \textbf{Det.} 
& \textbf{Spec.} 
& \textbf{Sent.} 
& \textbf{NetZ.} 
& \textbf{TCFD} 
& \textbf{WFB} 
& \textbf{WXI} 
& \textbf{Avg.} \\
\midrule

\rowcolor{ctxgray}
\multicolumn{11}{l}{\textit{LRD-specialized merging ($\theta_{\mathcal{X}}^{\textsc{lrd}}$)}} \\

$\theta_{\textsc{Soup}}$
& $83.6_{\pm1.6}$
& $66.5_{\pm3.7}$
& $95.4_{\pm0.6}$
& $70.9_{\pm2.1}$
& $76.7_{\pm1.4}$
& $98.9_{\pm0.0}$
& $\mathbf{64.1}_{\pm2.2}$
& $\mathbf{97.3}_{\pm0.1}$
& $\mathbf{33.7}_{\pm5.3}$
& $\mathbf{76.3}_{\pm1.9}$ \\

$\theta_{\textsc{TA}(0.5)}$
& $80.9_{\pm5.7}$
& $67.5_{\pm1.8}$
& $94.7_{\pm0.4}$
& $67.9_{\pm0.2}$
& $71.0_{\pm0.2}$
& $99.1_{\pm0.0}$
& $60.3_{\pm2.4}$
& $95.8_{\pm0.5}$
& $24.8_{\pm6.0}$
& $73.6_{\pm1.9}$ \\

$\theta_{\textsc{Norm}}$
& $84.3_{\pm1.8}$
& $67.3_{\pm1.7}$
& $94.1_{\pm1.1}$
& $67.5_{\pm1.0}$
& $78.0_{\pm1.5}$
& $98.2_{\pm0.1}$
& $61.4_{\pm0.7}$
& $96.4_{\pm0.3}$
& $23.8_{\pm3.4}$
& $74.6_{\pm1.3}$ \\

\midrule

\rowcolor{ctxgray}
\multicolumn{11}{l}{\textit{CX-only merging ($\theta_{\mathcal{X}}$) -- Ablation}} \\

$\theta_{\textsc{Soup}}^{\textsc{CX}}$
& $84.5_{\pm1.6}$
& $66.5_{\pm2.6}$
& $94.1_{\pm0.3}$
& $69.6_{\pm0.7}$
& $78.2_{\pm2.8}$
& $98.3_{\pm0.1}$
& $61.3_{\pm1.5}$
& $96.5_{\pm0.2}$
& $22.5_{\pm8.3}$
& $74.2_{\pm1.5}$ \\

$\theta_{\textsc{TA}(0.5)}^{\textsc{CX}}$
& $86.4_{\pm0.8}$
& $67.1_{\pm1.6}$
& $94.7_{\pm1.1}$
& $68.2_{\pm1.0}$
& $78.6_{\pm3.1}$
& $98.1_{\pm0.1}$
& $61.0_{\pm1.0}$
& $96.3_{\pm0.2}$
& $20.9_{\pm7.1}$
& $74.6_{\pm1.8}$ \\

$\theta_{\textsc{Norm}}^{\textsc{CX}}$
& $86.5_{\pm0.4}$
& $70.1_{\pm1.7}$
& $94.4_{\pm1.1}$
& $68.8_{\pm1.0}$
& $77.8_{\pm1.5}$
& $98.3_{\pm0.5}$
& $61.2_{\pm0.7}$
& $96.5_{\pm0.5}$
& $26.8_{\pm2.5}$
& $75.9_{\pm1.2}$ \\

\bottomrule
\end{tabular}}
\caption{
Model merging comparison between LRD-specialized and context-extension-only (CX-only) checkpoints.
All values report F$_1$ scores (\%, mean $\pm$ standard deviation across three random seeds).
LRD-specialized merging represents the primary adaptation setting following the ModernBERT training procedure, while CX-only merging is reported as an ablation to investigate whether effective parameter-space merging depends on LRD specialization.
}
\label{tab:cx_merging}
\end{table*}

\section{Training Setup: Full Configuration}
\label{app:training-setup}

Table~\ref{tab:full-config} reports the complete hyperparameter and architecture specification for every continued-pretraining run. All variants differ only in the data subset (see Table~\ref{tab:data_combinations}) and the learning-rate schedule. Final Composer-format checkpoints are converted to the Hugging Face Transformers \emph{FlexBERT} implementation for release.

\section{Context Extension-only Model Merging Analysis}
\label{app:cx-merging}

To further analyze the effect of the LRD specialization stage on model merging, we additionally evaluate merging strategies using context-extension-only checkpoints ($\theta_{\mathcal{X}}$). Unlike the main experiments, which use final CX+LRD checkpoints following the ModernBERT training procedure, this analysis isolates whether the observed benefits of parameter-space merging depend on the final LRD adaptation stage. We evaluate three merging strategies: simple weight averaging
($\theta_{\textsc{Soup}}$), Task Arithmetic with scaling factor
$\lambda=0.5$ ($\theta_{\textsc{TA}(0.5)}$), and norm-balanced
merging ($\theta_{\textsc{Norm}}$). Results are shown in Table~\ref{tab:cx_merging}.

Overall, CX-only merging achieves competitive performance across climate NLP benchmarks, suggesting that the effectiveness of model merging is not exclusively dependent on LRD specialization. However, the LRD-specialized checkpoints remain our primary setting because they represent the final adapted models following the established ModernBERT training pipeline.

\section{Comparison Against the Post-LRD Public ModernBERT Checkpoint}
\label{app:modernbert-full}

All climate-adapted variants in the main paper are initialized from the pre-LRD stable-phase ModernBERT-base checkpoint, denoted $\theta_{\textsc{base}}$. For completeness, we also fine-tune the final publicly released ModernBERT-base checkpoint --- the post-LRD release that downstream practitioners typically use --- on the same nine climate benchmarks under the identical fine-tuning protocol described in \S\ref{sec:tasks}. Table~\ref{tab:modernbert-full} reports per-task F$_1$ scores side-by-side with $\theta_{\textsc{base}}$.

The post-LRD checkpoint averages $72.7$ F$_1$, $0.8$ points below $\theta_{\textsc{base}}$ ($73.5$). Per-task, the largest regressions are on \textit{Retrieval} ($-3.6$), \textit{Commitments \& Actions} ($-5.3$), and \textit{TCFD} ($-2.2$); modest gains appear on \textit{Sentiment} ($+3.1$), \textit{WFB} ($+0.8$), \textit{NetZero} ($+0.2$), and \textit{Specificity} ($+0.1$); \textit{Detection} ($-0.1$) and \textit{WXImpactBench} ($-0.2$) are essentially unchanged.

\section{Downstream Tasks: Detailed Descriptions}
\label{app:tasks}

This appendix expands on the task list summarized in \S\ref{sec:tasks}. Table~\ref{tab:tasks} reports the type, number of classes, fine-tuning split sizes, and label semantics for each of the nine benchmarks; the paragraphs below describe each task's label space, the linguistic phenomenon it targets, and, where relevant, the construction pipeline and annotation provenance of the underlying dataset.

\paragraph{Single-label classification.}
\textbf{Climate Detection} \citep{bingler2024cheap} is a binary task that determines whether a passage is climate-related, requiring the model to distinguish domain-relevant content from superficially similar but topically unrelated text. \textbf{Climate Specificity} \citep{bingler2024cheap} classifies climate-related statements as either \textit{specific} or \textit{vague}, where specific statements contain concrete commitments or measurable targets; this distinction is central to assessing the substantiveness of corporate disclosures and to detecting greenwashing. \textbf{Climate Commitments and Actions} \citep{bingler2024cheap} categorizes text into forward-looking pledges versus descriptions of implemented measures, operationalizing the gap between corporate rhetoric and realized climate action. \textbf{Climate Sentiment} \citep{bingler2024cheap} assigns one of three labels --- \textit{risk}, \textit{opportunity}, or \textit{neutral} --- to climate-related paragraphs, capturing the evaluative stance toward climate change in corporate and policy discourse. \textbf{Net Zero and Reduction Targets} \citep{schimanski2023climatebert} classifies text into \textit{net-zero target}, \textit{reduction target}, or \textit{no target}, based on an expert-annotated dataset of $\sim$$3.5$K samples drawn from corporate reports and national pledges. \textbf{TCFD Recommendations} \citep{bingler2024cheap} assigns text to the disclosure categories of the Task Force on Climate-related Financial Disclosures, covering governance, strategy, risk management, and metrics and targets; the multi-class formulation requires the model to differentiate structurally similar but thematically distinct disclosure dimensions.

\paragraph{Multi-label classification.}
\textbf{Water, Forest, Biodiversity, and Nature (WFB Nature)} \citep{schimanski2023exploring} is grounded in the Taskforce on Nature-related Financial Disclosures framework, where each text sample may be labeled across four nature dimensions --- water, forest, biodiversity, and a general nature category --- requiring the model to capture co-occurring thematic content in corporate environmental disclosures. \textbf{WXImpactBench} \citep{yu2025wximpactbench} evaluates understanding of disruptive weather impacts on society across six impact categories: infrastructural, political, financial, ecological, agricultural, and human health. The dataset is constructed from historical newspaper articles through a four-stage pipeline of keyword extraction, event categorization, topic-aware article selection, and expert annotation.

\paragraph{Retrieval.}
\textbf{ClimRetrieve} \citep{schimanski2024climretrieve} benchmarks information retrieval from corporate climate disclosures. The dataset pairs $16$ detailed climate-related questions with $30$ sustainability reports, yielding over $8.5$K question-passage pairs annotated at multiple relevance levels. We follow the report-level setup of \citet{schimanski2024climretrieve} and cast the task as binary relevance classification by collapsing the original ordinal relevance scale at threshold $\geq 1$, so a single architecture and metric are shared with the classification tasks; the split sizes in Table~\ref{tab:tasks} count $(\text{query}, \text{passage}, \text{relevance})$ triples after the $80/10/10$ stratified split.

\end{document}